\documentclass[12pt]{article}

\usepackage{scicite}

\usepackage{times}

\newenvironment{sciabstract}{%
\begin{quote} \bf}
{\end{quote}}
\usepackage{graphicx,xcolor}
\usepackage[utf8]{inputenc}
\usepackage{caption}
\usepackage{subcaption}
\usepackage{amsmath,amsfonts,amssymb}
\usepackage[overload]{empheq} 
\usepackage{soul}
\usepackage{bm}
\usepackage{multirow}
\usepackage{hyperref}
\usepackage{xcolor}
\usepackage{authblk}
\usepackage{lineno}
\linenumbers
\usepackage{setspace}
\usepackage[english]{babel}
\usepackage[backend=biber,style=science,articletitle=true,url=false]{biblatex}
    \title{Fine-grained prediction of food insecurity using news streams}

\author[1]{Ananth Balashankar}
\author[1,2]{Lakshminarayanan Subramanian}
\author[1,3,4,*]{Samuel P. Fraiberger}
\affil[1]{Department of Computer Science, New York University, New York, NY}
\affil[2]{Department of Population Health, NYU School of Medicine, New York, NY}
\affil[3]{Development Data Group, World Bank, Washington, DC}
\affil[4]{Connection Science, Massachussets Institute of Technology, Cambridge, MA}
\affil[*]{sfraiberger@worldbank.org}

\begin{document}
\nolinenumbers
\maketitle

\begin{sciabstract}
Anticipating the outbreak of a food crisis is crucial to efficiently allocate emergency relief and reduce human suffering. However, existing food insecurity early warning systems rely on risk measures that are often delayed, outdated, or incomplete. Here, we leverage recent advances in deep learning to extract high-frequency precursors to food crises from the text of a large corpus of news articles about fragile states published between 1980 and 2020. Our text features are causally grounded, interpretable, validated by existing data, and allow us to predict 32\% more food crises than existing models up to three months ahead of time at the district level across 15 fragile states. These results could have profound implications on how humanitarian aid gets allocated and open new avenues for machine learning to improve decision making in data-scarce environments.
\end{sciabstract}

\newpage
\noindent Food insecurity continues to threaten the lives of hundreds of millions of people around the world today. According to the Food and Agriculture Organization of the United Nations (FAO), the number of undernourished increased from 624 million people in 2014 to 688 million in 2019 \cite{fao}. There is considerable evidence that quickly responding to emerging risks of food insecurity saves lives and lowers humanitarian costs \cite{ravallion}, which leads aid agencies to resort to early warning systems to decide when and where to deploy emergency relief \cite{backer}. While risk factors are well-established \cite{sen,mellor,maxwell,yared}, ranging from conflicts to pests, weather shocks and migration, delayed or infrequent measurements of these factors typically impede early warning systems' ability to promptly anticipate food crises \cite{bo2}. Furthermore, fragile states prone to food insecurity often lack the capacity to systematically measure risk factors, generating data gaps \cite{WDR}. Against this backdrop, the past decade has seen an explosion in the availability of vast repositories of digital data, from satellite imagery to call detailed records, which are increasingly being analyzed to address socioeconomic challenges \cite{jean,blumenstock}. Encouraged by these approaches, we take advantage of recent advances in deep learning and natural language processing to extract anticipatory signals of food insecurity episodes from the text of a large corpus of news articles. Unlike existing food insecurity early warning systems, the articles we collect are published on a daily basis, allowing us to generate high-frequency forecasts \cite{bo}. News aggregators provide access to media articles curated in a transparent manner going back several decades, enabling the analysis of long time series of news streams \cite{wang}. Finally, authoritative media sources such as \textit{BBC} or \textit{Reuters} have a long-standing reputation for providing trustworthy information on local contexts, suited to produce disaggregated forecasts \cite{bhatia}. 

\begin{figure}[!ht]
\centering
\includegraphics[width=\linewidth]{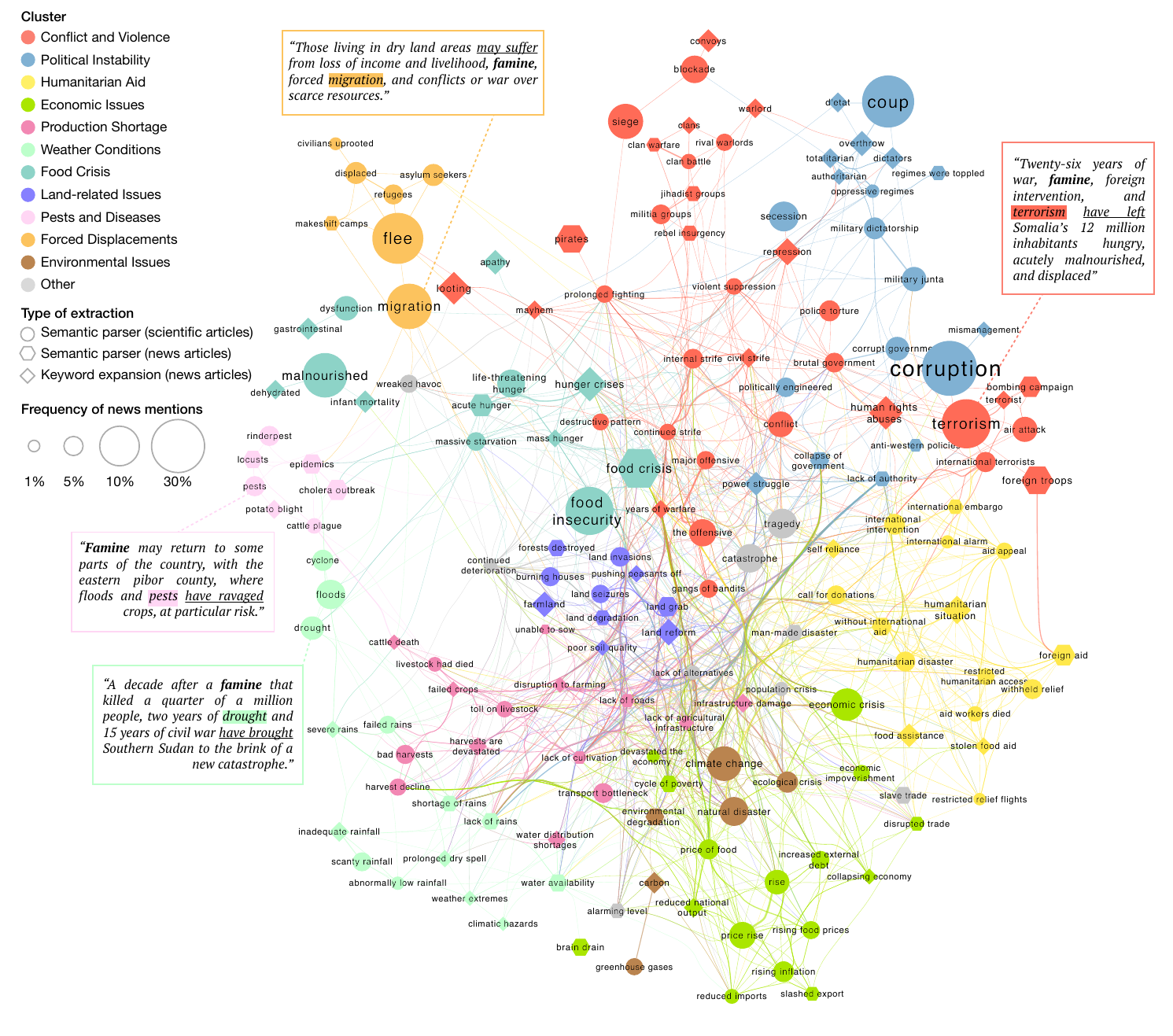}
\caption{\textbf{Uncovering mentions of causes of food insecurity.} \footnotesize{Starting from a handcrafted list of 13 target keywords related to food insecurity, we use a frame-semantic parser to extract from scientific (circles) and news (hexagons) articles a seed list of causes of food insecurity (“text features”) mentioned in the same semantic frame as a target keyword \cite{baker,swayamdipta}. Each box contains an example of a sentence in which the parser detects a text feature (highlighted in color) mentioned in the same semantic frame as the target keyword “\textit{famine}” (in bold) and a causal link (underlined). We expand this seed list by collecting text features from news articles (diamond) that are semantically similar to a seed according to their word mover's distance \cite{mikolov,kusner}. Text features for which the proportion of monthly local news mentions fails to predict the IPC classification of food security are discarded, leading to a final set of 167 features grouped into 12 clusters based on their semantic similarity and mapped onto a network. A node's size is proportional to its text feature's frequency in news articles mentioning target keywords, and an edge's width encodes the semantic proximity between its end nodes text features. A force-directed algorithm determines each node's position, leading nodes representing semantically similar text features to appear close to one another.}}
\label{fig1}
\end{figure}

\begin{figure}[!ht]
\centering
\includegraphics[width=\linewidth]{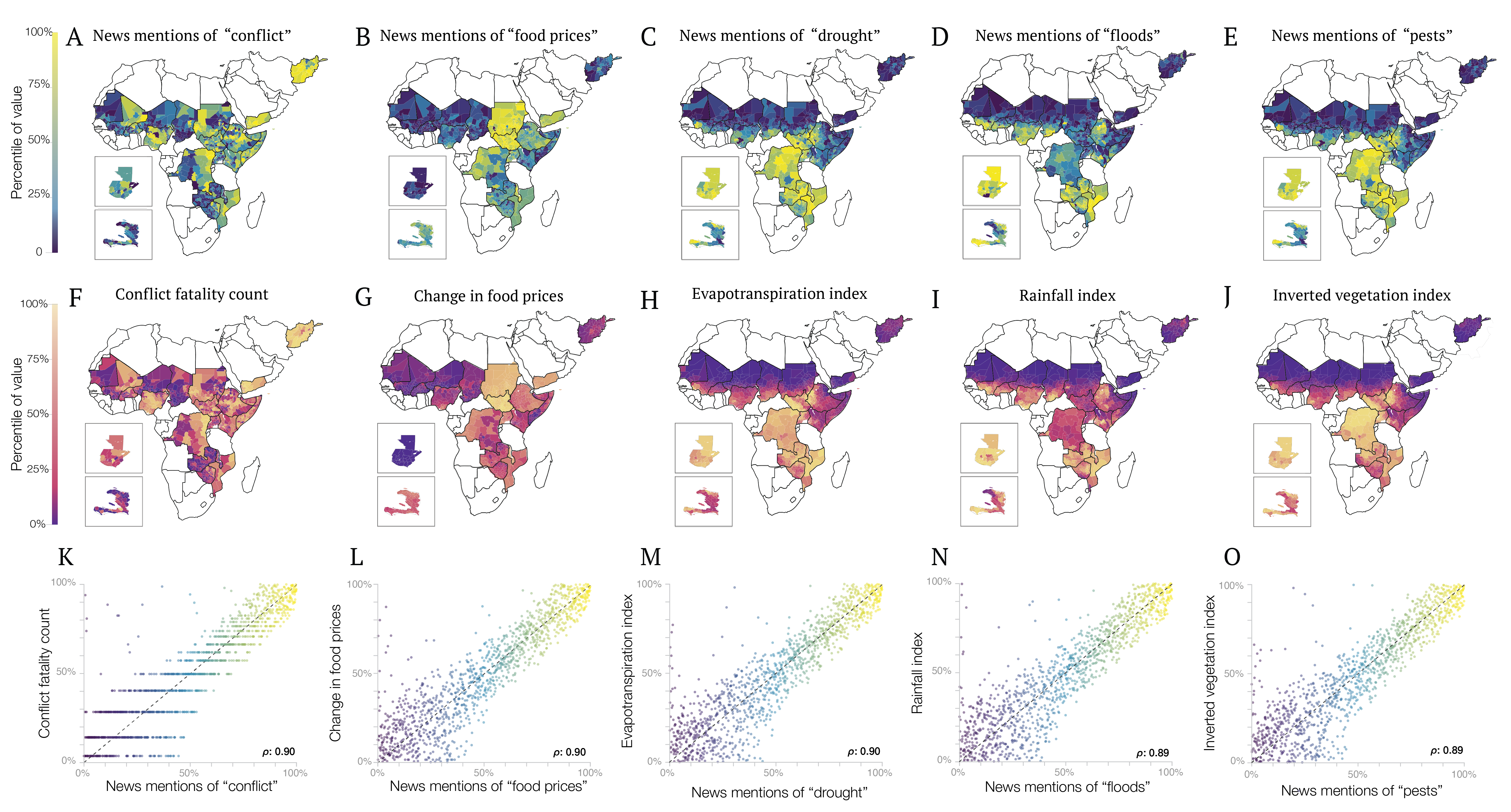}
\caption{\textbf{Validating news-based indicators of food insecurity.} \footnotesize{We demonstrate that there exists a strong cross-sectional relationship between news mentions (A-E) and traditional measures (F-J) of causes of food insecurity. We use a comprehensive dataset of traditional measures of food insecurity risk factors (“traditional factors”) across 21 fragile states during the period 2009-2020 -- a conflict fatality count, the change in food prices, an evapotranspiration index, a rainfall index, and an inverted vegetation index -- summarizing each district by the maximum monthly value of each traditional factor during the observation period. To uncover the text feature most closely related to each traditional factor, we first summarize each district by the maximum monthly proportion of local news articles mentioning each text feature (“news factor”). We then associate with each traditional factor the news factor with which it has the highest Spearman correlation across district. (K-O) A scatter plot of a traditional factor (y-axis) and its associated news factor (x-axis) across districts reveals a high Spearman correlation ($r_S>0.89$). All the values are reported in percentiles.}}
\label{fig2}
\end{figure}

Our study focuses on predicting the Integrated Phase Classification (IPC) of food insecurity. This classification is available at the district level across 37 fragile states in Africa, Asia, and Latin America, and was reported every four months between 2009 and 2015 and every three months thereafter. Food insecurity is classified according to an ordinal scale composed of five phases: minimal, stressed, crisis, emergency, and famine (supplementary text \ref{IPC}). We postulate that the factors triggering a food crisis are mentioned in the news prior to being potentially measured by traditional risk indicators. We therefore collect a novel dataset from the news aggregator \hyperlink{https://professional.dowjones.com/factiva/}{Factiva} containing 11.2 million articles covering countries included in the IPC dataset and published between June 1980 and July 2020 (supplementary text \ref{news}). We then develop a methodology based on semantic role labeling to elicit textual mentions of causes of food insecurity \cite{baker,swayamdipta}. We start from a handpicked list of 13 target keywords related to food insecurity and we use a frame-semantic parser to uncover causes of food insecurity appearing in the same semantic frame as one of the target keywords (supplementary text \ref{frame_semantic_parsing}). For example, when the parser examines the sentence “\textit{Famine may return to some parts of the country, with the eastern Pibor county, where floods and pests have ravaged crops, at particular risk}”, it detects that “\textit{floods}” and “\textit{pests}”, both being established causes of food insecurity, are mentioned in the same semantic frame as the target keyword “\textit{famine}”. We apply this method on our corpus of news which allows us to elicit 1,062 text features consisting of unigrams, bigrams and trigrams occurring in the same semantic frame as a target keyword. To ensure that we capture a wide range of causes of food insecurity recognized both by journalists and by experts, we repeat the same procedure on a manually selected list of 93 peer-reviewed studies and books on food insecurity, which reveals 149 additional text features. We then expand this seed list of 1,211 features by considering text features mentioned in the news which are semantically similar to a seed \cite{mikolov,kusner}, obtaining 738 new features (supplementary text \ref{keyword_expansion}). Finally, to drop any irrelevant text feature that might have accidentally been picked up, we first convert each text feature into an index per month and per district by computing the proportion of monthly news articles mentioning both the text feature and the district (“news factor”). We then discard news factors which are not predictive of the IPC phase, leading to a final set of 167 text features (supplementary text \ref{dimensionality_reduction}). To shed light on these features, we partition them into 12 semantically distinct clusters (Fig. \ref{fig1}). We find that text features belonging to the same clusters tend to co-occur in the news, the average pairwise correlation between news factors in the same cluster being 69.9\% versus 34.9\% for those in different clusters, which provides support to our partitioning (Fig. \ref{fig8}). We also find that 9 of 12 clusters are composed of text features related to known causes of food insecurity -- “conflict and violence”, “political instability”, “economic issues”, “production shortage”, “weather conditions”, “land-related issues”, “pests and diseases”, “forced displacements”, and “environmental issues” -- accounting for 92\% of the articles in which text features are mentioned. The remaining 3 clusters include terms related to “food crises”, to “humanitarian aid”, and “other” negative terms unspecific to food insecurity. Having established the consistency between our text features and known causes of food insecurity, we also demonstrate the presence of a strong cross-sectional relationship between news mentions and traditional measures of causes of food insecurity (Fig. \ref{fig2}). We use a comprehensive dataset containing traditional measures of food insecurity risk factors (“traditional factors”) used in early-warning systems -- a conflict fatality count, the change in food prices, an evapotranspiration index, a rainfall index, and an inverted vegetation index -- covering 21 fragile states over the period 2009-2020 (supplementary text \ref{risk_factors}). After summarizing each traditional factor and each news factor at the district level with its maximum monthly value during the observation period, we associate with each traditional factor the news factor with which it has the highest Spearman correlation across districts (Fig. \ref{fig2}A-\ref{fig2}J). We find that the conflict fatality count, change in food prices, evapotranspiration index, rainfall index, and inverted vegetation index are most strongly correlated to news mentions of “conflict”, “food prices”, “drought”, “floods”, and “pests” respectively ($r_S>0.89$), thereby providing an additional sanity check for our approach (Fig. \ref{fig2}K-\ref{fig2}O). Taken together, these results indicate that our procedure allows us to uncover text features that are consistent with established causes of food insecurity, interpretable, and validated by traditional risk indicators.

\begin{figure}[!ht]
\centering
\includegraphics[width=\linewidth]{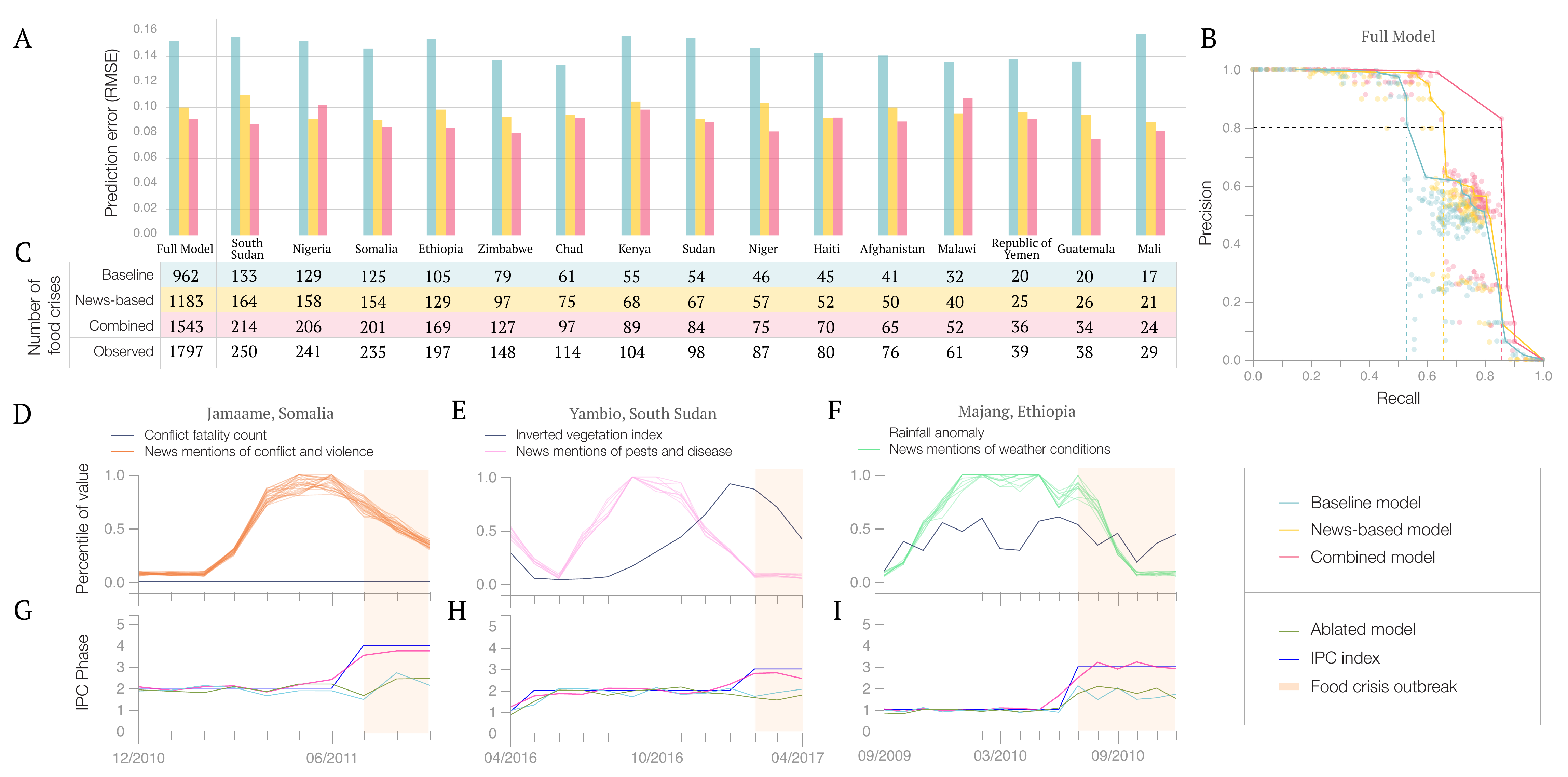}
\caption{\textbf{Predicting food insecurity.} \footnotesize{(A) We show that the monthly proportion of local news articles mentioning a text feature (“news factor”) helps predict the IPC classification of food security at the district level across 15 fragile states during the period 2009-2020. We estimate a panel autoregressive distributed lag model of the IPC phase on past values of traditional risk factors (“baseline model”, turquoise bars), news factors (“news-based model”, yellow bars), and both sets of factors (“combined model”, pink bars). We report the average root-mean-square error (y-axis) over 10 cross-validated periods, which reveals that including news factors leads to an average reduction in prediction error of 40\% relative to the baseline model, with gains ranging from 20.5\% for Malawi to 48.4\% for Mali. (B) We turn each previously estimated model into a classifier of the outbreak of a food crisis, characterized by the IPC phase raising to a value of 3 or more for at least two consecutive periods. By varying the classification threshold, we construct a series of classifiers (dots) with different precision (y-axis) and recall (x-axis), allowing us to uncover each model's Pareto front (full lines). We then choose each model's threshold such that its precision is equal to 80\% (black dotted line), finding that the combined model's recall reaches 86\%, compared to 66\% for the news-based model and 54\% for the baseline model (colored dotted lines). (C) Number of crisis outbreaks observed in the validation set (white row) and predicted by the baseline (turquoise row), news-based (yellow row) and combined (pink row) model at a fixed precision of 80\%. (D-F) To elicit the role played by news factors in driving our predictions, we zoom in on 3 crisis episodes in the validation set during which news mentions of causes of food insecurity included in the “conflict and violence” (orange lines), “pests and diseases” (pink lines), and “weather conditions” cluster (green lines) would have helped anticipate the deterioration of the situation. For each episode, we report each text feature's proportion of monthly local news mentions and the most closely related traditional risk indicator (black line). All the values are reported in percentiles. (G-I) We also report the time series of the IPC phase (blue line), and its predicted value using the baseline (turquoise line), ablated (khaki line), and combined (red line) model. While risk factors measured with traditional data fail to provide a warning signal in a timely fashion, news factors peak prior to each crisis outbreak (mauve shaded area), leading the combined model to accurately predict the outbreak whereas the baseline and ablated models fail to predict it.}}
\label{fig3}
\end{figure}

Next, we demonstrate that news factors help predict variations in food insecurity in fragile states (Fig. \ref{fig3}). Following previous research on food insecurity early warning systems \cite{lentz2019data,bo2}, we first estimate a panel autoregressive distributed lag (ADL) model to predict the IPC phase using past values of the traditional factors described in Fig. \ref{fig2} along with time-invariant risk factors -- population count, district size, terrain ruggedness and agricultural land use share (“baseline model”). We then compare the baseline model's predictive performance to that of the same ADL model in which we substitute traditional factors with news factors (“news-based model”), finding that the news-based model leads to a reduction in out-of-sample root-mean-square error (RMSE) of 34.1\%  relative to the baseline model (Fig. \ref{fig3}A). These results suggest that news factors could serve as a proxy for food insecurity risk factors in regions in which other sources of data are unavailable or outdated. Furthermore, when we incorporate both traditional factors and news factors into the same ADL model (“combined model”), the reduction in RMSE relative to the baseline model reaches 40\%, proving that news-based indicators of food insecurity also serve as a complement to traditional risk indicators (supplementary text \ref{regression_model}). While these results show that news factors improve the prediction of variations in food security at the district level, we find a substantial degree of heterogeneity in predictive gains across countries, ranging from 20.5\% for Malawi to 48.4\% for Mali, which is in part explained by differences in news coverage (Fig. \ref{fig10}). To put these results into perspective, we also demonstrate that news factors specifically help predict the outbreak of a food crisis, which corresponds to the IPC phase raising to a value of 3 or more for at least two consecutive periods, an event of utmost importance to disaster relief organizations deciding when and where to allocate emergency food assistance (supplementary text \ref{food_crises_classification}). By converting each previously estimated model into a binary classifier of a crisis outbreak and fixing its precision at 80\%, we find that the combined model's out-of-sample recall reaches 86\%, compared to 66\% for the news-based model and 54\% for the baseline model (Fig. \ref{fig3}B). In other words, while the baseline model is able to predict 962 out of the 1,797 food crises observed in our validation set, incorporating news factors helps anticipate 581 additional crises which would have otherwise been missed (Fig. \ref{fig3}C). In addition, the combined model predicts 47 out of the 48 crisis outbreaks in which the IPC phase escalated to a level 4 or 5, while the baseline and news-based model only predict 26 and 33 of these outbreaks respectively, indicating that news factors are especially valuable in anticipating the most severe outbreaks. Taken together, these results show that news mentions of causes of food insecurity precede variations in district-level IPC phases and could help dispatch emergency relief up to three months ahead of a food crisis.

While machine learning is often criticized for its lack of transparency \cite{lipton}, our model's predictions can easily be interpreted. Focusing on Somalia, South Sudan and Ethiopia, three of the countries which experienced the highest level of food insecurity in recent decades, we zoom in on specific crisis episodes in our validation set to elicit which news factors help predict the deterioration of the situation. The first episode that we analyze happened in 2011 in Somalia, where the combination of a drought, rising food prices, forced displacement and a sustained conflict led to the worst famine of the 21st century \cite{somalia}. In particular, the district of Jamaame evolved from an IPC phase 2 during the first half of 2011 to a phase 4 by July, following intensifying violence in the Southern part of the country. While the proportion of news articles mentioning both Jamaame and terms included in the “conflict and violence” cluster started raising 5 months prior to the change in the IPC phase, the conflict fatality count did not record any death in the district until the summer of 2012, highlighting that news factors capture relevant dimensions of civil insecurity which are missing from traditional conflict indicators (Fig. \ref{fig3}D and \ref{fig3}G). The second episode that we focus on occurred in 2016 when a fall armyworm spread across 20 countries in Africa, decimating large quantities of crops \cite{southsudan}. The worm was first reported in early 2016, and by September, the proportion of news mentioning both the Yambio county in South Sudan and text features included in the “pests and diseases” cluster had peaked, 4 months prior to the inverted vegetation index peaking, and 5 months ahead of the IPC phase raising from 2 to 3 (Fig. \ref{fig3}E and \ref{fig3}H). Although pest infestations are indirectly measured through vegetation indices, their damage on crops are typically only reflected in vegetation greenness once the food security of neighboring populations has begun to deteriorate, strengthening the importance of measuring anticipatory signals from the news. Finally, the last episode of our study took place in 2009 in Ethiopia when it experienced one of its driest years of the past 50 years, wreaking havoc on food production \cite{ethiopia}. Seasonality-adjusted levels of precipitation in the Majang district were 2.3 standard deviations below their historical average in September 2009 before reverting to their mean at the beginning of 2010. While the prolonged effect of this extreme drought was not well captured by a precipitation index, the proportion of news mentioning both Majang and terms contained in the “weather conditions” cluster started increasing in late 2009 and remained close to its peak until July of 2010 when the IPC phase increased from 1 to 3, suggesting that news indices are also better suited to anticipate a drought-related food crisis (Fig. \ref{fig3}F and \ref{fig3}I). To quantify the role played by news factors in driving our predictions during these episodes, we re-estimate the combined model after having removed the cluster of news factors containing terms related to “conflicts and violence”, “pests and diseases”, and “weather conditions” respectively (“ablated models”). For all 3 episodes, we find that the combined model is able to accurately anticipate the change in the IPC phase whereas both the ablated model and the baseline model fail to predict it (Fig. \ref{fig3}G-\ref{fig3}I and \ref{fig12}). In other words, risk factors leading to a food crisis are better anticipated by news indicators than by traditional ones which can be incomplete, delayed or outdated, and our model enables us to explicitly interpret predictions of food crisis outbreaks by linking them to variations in news mentions of the underlying causes of an upcoming outbreak.

Although the drivers of food insecurity are well-known, early warning systems relying on high-frequency measurements of these factors are still lacking. The data-driven approach described in this paper could drastically improve the prediction of food crisis outbreaks up to three months ahead of time using real-time news streams and a predictive model that is simple to interpret and explain to policymakers. Development practitioners working for humanitarian organizations such as the World Food Program could use the predictions of our model to help prioritize the allocation of emergency food assistance across vulnerable regions in a principled way, allowing for a more effective preparedness and a reduction in human suffering when a crisis hits. Early warnings cannot address all of the sources of delay in emergency responses, however it can mitigate it by increasing the cost of inaction for governments and the international community \cite{dempsey}. While our study only focuses on news articles in English, future work incorporating local languages into our framework could potentially improve the predictive performance of our model even further. In addition, development practitioners could extend our model to produce estimates of the IPC phase during periods or in regions in which it is not currently being reported, at a fraction of the cost. Beyond the context of food insecurity, our novel approach for selecting causally grounded news indicators addresses the risk of overfitting when big data and machine learning is being used to predict policy outcomes in data-scarce environments \cite{wang,lazer,kleinberg}, and could be extended to other domains, from disease surveillance to the impact of climate change.

\printbibliography

\section*{Figures}

\clearpage
\renewcommand{\thefigure}{S\arabic{figure}}
\renewcommand{\thesection}{S\arabic{section}}  
\renewcommand{\thetable}{S\arabic{table}}
\setcounter{figure}{0}
\setcounter{table}{0}

\section*{Methods}
\section{Data collection} 
\subsection{Food insecurity classification data}
\label{IPC}
\begin{figure}[!ht]
\centering
\includegraphics[width=\linewidth]{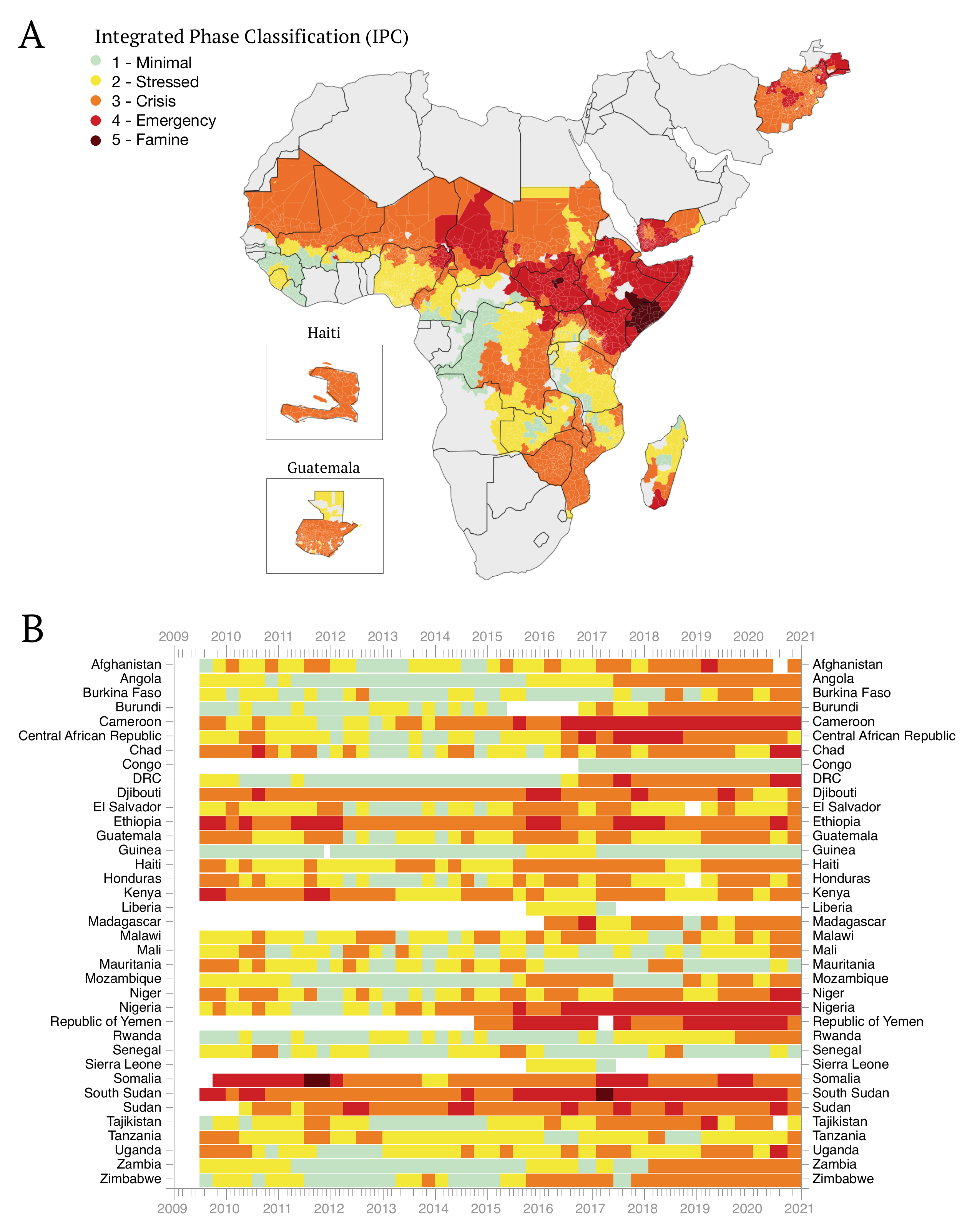}
\caption{\textbf{Food security dataset.} \footnotesize{(A) Integrated Phase Classification (IPC) of food security into 5 phases -- minimal, stressed, crisis, emergency, and famine -- at the district level across the 37 countries covered by the FEWS NET dataset. Each administrative unit is characterized by the maximum IPC phase measured over the period 2009-2020, revealing that food insecurity is geographically clustered. (B) Heatmap showing the maximum value of the IPC phase at the country level during each measurement period.}}
\label{fig4}
\end{figure}

Our dataset on food insecurity comes from the Famine Early Warning System Network (\hyperlink{https://fews.net/}{FEWS NET}). Food insecurity is classified into 5 phases following the Integrated Phase Classification (IPC) framework: (1) minimal, (2) stressed, (3) crisis, (4) emergency, and (5) famine. Phases are determined by experts and published at the district level across 37 countries since 2009, allowing us to compare food insecurity levels across time and regions in a standardized way. The classification covers the following countries: Afghanistan, Angola, Burundi, Central African Republic, Cameroon, Chad, Congo, El Salvador, Ethiopia, Guatemala, Guinea, Haiti, Honduras, Kenya, Liberia, Madagascar, Malawi, Mali, Mauritania, Mozambique, Nigeria, Niger, Rwanda, Senegal Sierra Leone, Somalia, South Sudan, Sudan, Tajikistan, Djibouti, Tanzania, Uganda, Burkina Faso, Republic of Yemen, Democratic Republic of the Congo, Zambia, and Zimbabwe. It was established 4 times per year from 2009 to 2015 -- in January, April, July, October -- and 3 times per year thereafter -- in February, June, October (Fig. \ref{fig4}).

\subsection{Corpus of news articles} 
\label{news} 
\begin{figure}[!ht]
\centering
\includegraphics[width=\linewidth]{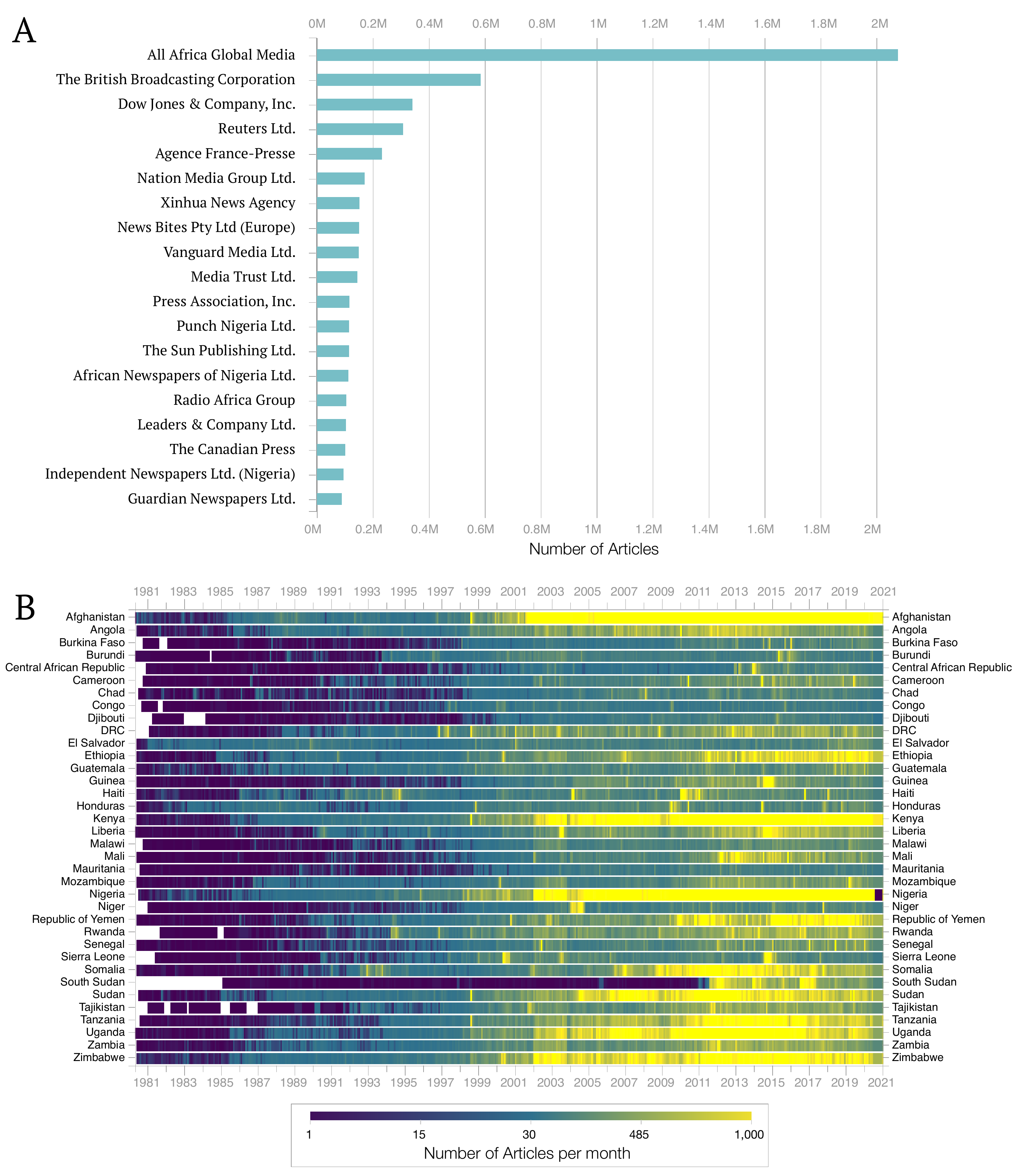}
\caption{\textbf{Corpus of news articles.} \footnotesize{(A) The number of news articles grouped by publisher. (B) The number of news articles grouped by month and by country. We use the classification provided by Factiva to establish that an article focuses on a specific country.}}
\label{fig5}
\end{figure}

Our dataset of news articles comes from Factiva, a digital archive of global news content which aggregates more than 33,000 news resources from 200 countries in 28 languages. Each news article is tagged with geographic region codes, allowing us to ascertain its relevance to a specific country. We collect the text of the 11.2 million articles in English tagged with at least one of the 37 countries covered by FEWS NET (Fig. \ref{fig5}). While 60.5\% of the articles were published by news sources located in a fragile state, the remaining 39.5\% come from news sources based in a non-fragile state.

\section{Feature selection}
\label{feature_selection}
\subsection{Frame-semantic parsing}
\label{frame_semantic_parsing}
\begin{figure}[!ht]
\centering
\includegraphics[width=\linewidth]{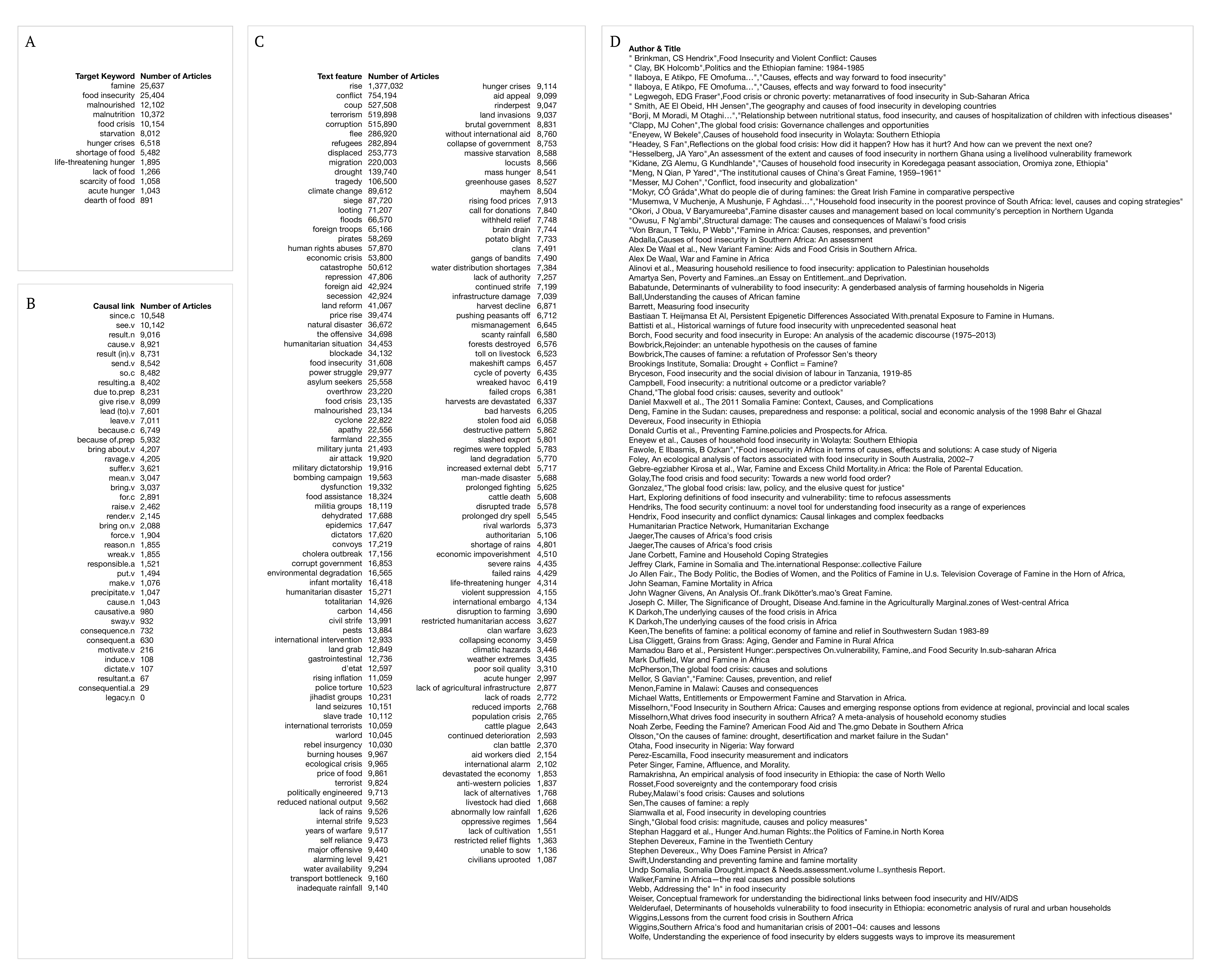}
\caption{\textbf{Semantic-frame parsing.} \footnotesize{(A) The 13 target keywords used to select semantic frames related to food insecurity along with the number of news articles in which the selected frames appear. To account for possible inflections, we use the Porter stemming algorithm on each word token and we select from our news corpus semantic frames matching the root words. (B) The 41 causal links obtained from the FrameNet lexical database used to select relevant semantic frames, along with the number of news articles in which the selected frames appear. (C) The 167 text features used in our predictive model along with the number of news articles in which they appear. (D) The 93 books and peer-reviewed studies on which we also run our semantic parser.}}
\label{fig6}
\end{figure}

We use a frame-semantic parser to extract from our corpus of news articles the text features which are causally related to food insecurity  \cite{baker}. The parser first splits each sentence into syntactic constituents $c_1, c_2,...,c_k$, where each $c_i$ includes $p \geq 0$ contiguous word tokens $w_j, w_{j+1},...,w_{j+p}$ starting from position $j$. It then assigns to each syntactic constituent $c_i$ a semantic role $t_i$. We use a deep neural network model reaching state-of-the-art accuracy on the benchmark dataset FrameNet to predict the semantic role of each syntactic constituent in our news corpus \cite{swayamdipta}. A semantic frame $f$ is a collection of syntactic constituents along with their semantic roles $(c_i,t_i)_{i \in f}$ . To select text features corresponding to causes of food insecurity, we restrict the set of semantic frames produced by the parser using the following filters:
\begin{enumerate}
   \item First, we exclude semantic frames whose constituents' roles do not include at least one “cause” and one “effect”. Note that there might more than one “cause” and one “effect” per frame.
    \item Next, we exclude semantic frames in which the “effect” constituents do not contain any of our 13 target keywords related to food insecurity (Fig. \ref{fig6}A).
    \item Next, we exclude semantic frames which do not contain any of the causal links included in the FrameNet lexical database (Fig. \ref{fig6}B). 
    \item Finally, we select all the unigrams, bigrams, and trigrams mentioned either in a “cause” or in an “effect” constituent of the previously selected frames (Fig. \ref{fig6}C). 
\end{enumerate}
This procedure allows us to elicit 1,062 text features from our corpus of news articles. To ensure that our text features cover the causes of food insecurity established by experts, we also handpick a list of 93 well-cited books and peer-reviewed studies on food insecurity from Google Scholar using the following queries: “causes of famine”, “food insecurity causes”, “food insecurity Africa causes”, “causes of food crisis”, “famine Africa”, “famine Africa causes” and “food crisis Africa causes”. We then run the parser on the text of these books and studies which reveals 149 additional text features (Fig. \ref{fig6}D).

\subsection{Keyword expansion}
\label{keyword_expansion}
\begin{figure}[!ht]
\centering
\includegraphics[width=\linewidth]{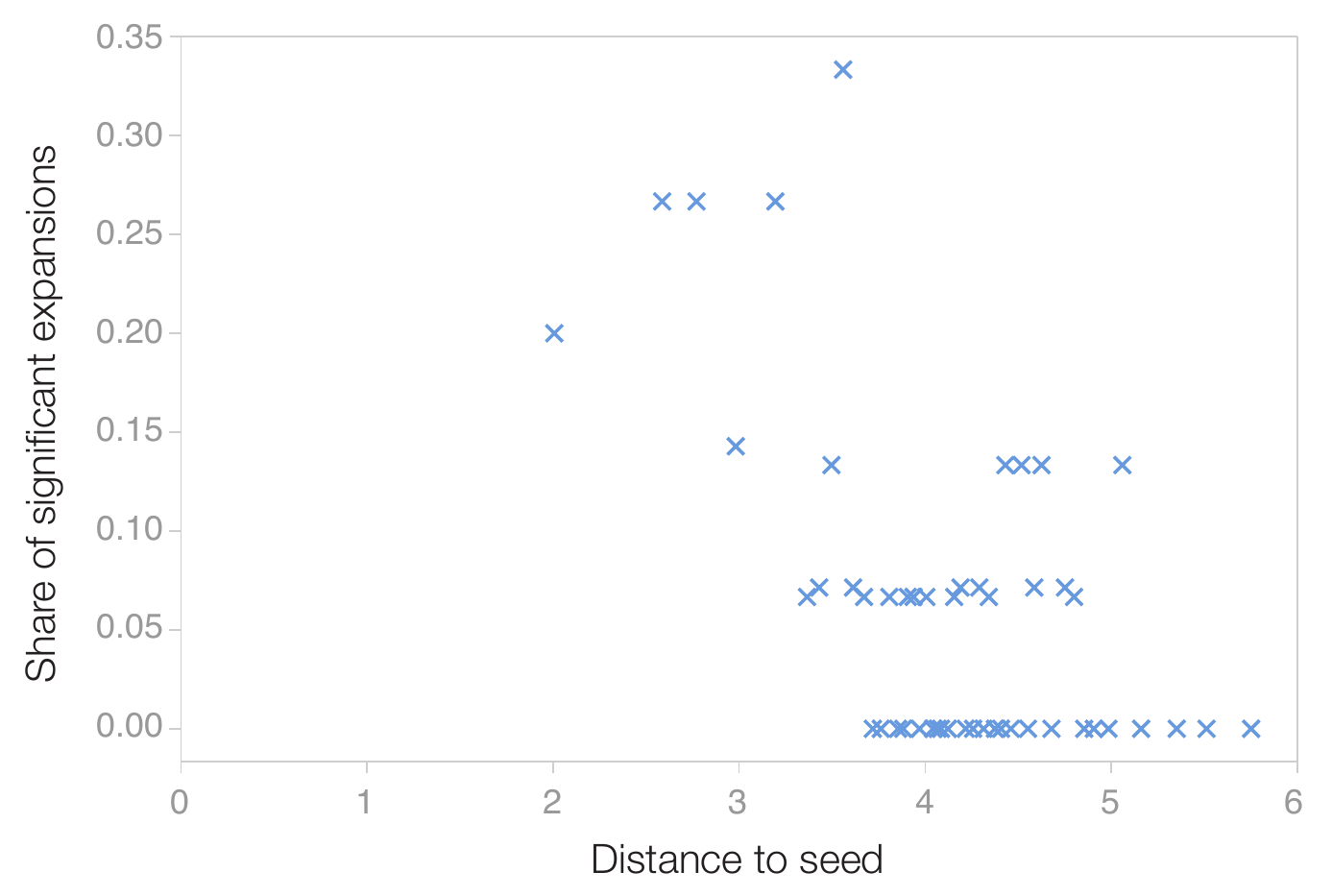}
\caption{\textbf{Keyword expansion.} \footnotesize{Starting from the 101 seeds obtained by semantic-frame parsing and passing the Granger causality test, we find the 738 candidate features mentioned in the news and with a word mover's distance to a seed smaller than 6. After ranking candidate features by increasing distance to a seed and partitioning them into 50 groups of equal size, we report the proportion of candidate features within each group passing the Granger causality test (y-axis) and the average distance to a seed within each group (x-axis). As the distance to a seed gets close to 6, the proportion of candidate features predicting the IPC phase approaches zero, providing support to our choice of exploring the space of semantic neighbors up to a distance of 6.}}
\label{fig7}
\end{figure}

While our semantic parser allows us to extract a seed list of 1,211 features causally linked to food insecurity, it fails to capture words semantically close to a seed that are also relevant. For example, the parser selects the word “terrorism” but not the equally relevant word “terrorist” which does not appear in any of the causal frames that we considered. For this reason, we expand our seed list of features with words and phrases semantically close to each seed. We do so by considering as candidate features all the unigrams in our news corpus as well as all the bigrams and trigrams occurring more than 1,000 times. We then convert the words of each feature into an embedding vector such that words occurring in similar contexts end up close to one another in the embedding space \cite{mikolov}. By computing the word mover's distance between each seed and each candidate feature, we keep the candidates whose distance to a seed is smaller than 6, obtaining 738 new features \cite{kusner}. We find that expanding to more distant candidates does not lead to additional relevant features, which indicates that the features that we select cover a wide range of causes of food insecurity (Fig. \ref{fig6}C and Fig. \ref{fig7}).

\subsection{Dimensionality reduction}
\label{dimensionality_reduction}
Having uncovered a set of 1,949 text features causally related to food insecurity, we now aim to focus on those whose news mentions help predict the IPC phase. We first convert each text feature $w$ into a time series $x_{w, d, t}$ measuring the proportion of news articles mentioning $w$ and the district $d$ during the month ending on date $t$ (“news factor”). We then convert the IPC phase into a monthly indicator by forward filling the latest observation. We then estimate a panel autoregressive distributed lag model of the IPC phase $y_{d,t}$ in district $d$ during the month ending on date $t$:

\begin{align}
    y_{d,t} = a_0 + a_1 y_{d,t-1} + a_2 y_{d,t-2} + ... + a_n y_{d,t-n} + b_1 x_{w, d, t-1} + ... + b_n x_{w, d, t-n},
\end{align}

\noindent where the number of lags $n$ is chosen based on the Akaike Information Criterion. This estimation is done at the district level which is the smallest geographic unit for which the IPC phases are observed. To determine whether a news factor predicts the IPC phase, we run a Granger-causality test and we reject the null hypothesis that $x_{w}$ does not Granger-cause $y$ if the news factor and its lagged values whose coefficients are statistically different from zero add explanatory power to the regression according to an F-test at the 1\% level. Since the Granger causality test assumes stationarity, we take the first difference of non-stationary time series until it passes the Augmented Dickey-Fuller test. If a news factor is predictive of the IPC phase after differentiation, we keep the transformed news factor for the rest of the analysis. Out of the 1,949 text features previously selected, we retain the 167 text features which are predictive of the IPC phase (Fig. \ref{fig6}C).

\section{Predicting food insecurity}
\subsection{Traditional risk indicators}
\label{risk_factors} 
To uncover the predictive value of our news factors, we compare them with traditional measures of food insecurity risk factors obtained from recent studies on food insecurity \cite{bo2,bo,bo3}. Our risk indicators include:
\begin{itemize}
    \item a monthly count of violent conflict events and the monthly average number of fatalities per event
    \item a food prices index (monthly log nominal food price index
and monthly year-on-year difference) 
    \item an evapotranspiration index (monthly mean)
    \item a rainfall index (monthly mean and deviation from average seasonal value)
    \item a normalized difference vegetation index (monthly mean and deviation from average seasonal value)
    \item a population count
    \item a terrain ruggedness index
    \item the district size
    \item the share of cropland use 
    \item the share of pasture use 
\end{itemize}
In other words, we collect district-level data on 9 time-varying risk indicators describing 5 different types of risk, and 5 time-invariant risk indicators. The dataset covers 21 out of the 37 countries covered by FEWS NET -- Afghanistan, Burkina Faso, Chad, Democratic Republic of the Congo, Ethiopia, Guatemala, Haiti, Kenya, Malawi, Mali, Mauritania, Mozambique, Niger, Nigeria, Somalia, South Sudan, Sudan, Uganda, Republic of Yemen, Zambia, and Zimbabwe. To build our predictive models, we focus on the subset of 15 countries experiencing more than 20 food crises during the observation period -- Afghanistan, Chad, Ethiopia, Guatemala, Haiti, Kenya, Malawi, Mali, Niger, Nigeria, Somalia, South Sudan, Sudan, Republic of Yemen, and Zimbabwe. This dataset contains 33,847 monthly observations across 915 districts between July 2009 and February 2020 (Fig. \ref{fig8}).

\subsection{Regression model}
\label{regression_model}
Let $v_{k,d,t}$ be the value of time-varying risk indicator $k$ in district $d$ during the month ending on date $t$ and let $v_{l,d}$ be the value of time-invariant risk indicator $l$ in district $d$. In addition, let $x_{w,d,t}$ be the news factor measuring the proportion of news articles mentioning text feature $w$ and district $d$ during the month ending on date $t$. To account for news mentions of causes of food insecurity co-occurring with the name of a province or a country, we also introduce $x_{w,p_d,t}$ and $x_{w,c_d,t}$ where $p_d$ and $c_d$ respectively correspond to the province and the country that district $d$ belongs to. Finally, let $y_{d,t}$ be the IPC phase in district $d$ during the quarter ending on date $t$, such that missing data are filled forward using the latest available data. To predict the IPC phase, we estimate the following panel autoregressive distributed lag (ADL) model:

\begin{align}
\label{equation_model}
    y_{d,t} &= a_{d} + \sum_{m=1}^{6} a_{d, m} y_{d, t-3m} + \sum_{k=1}^{9} \sum_{n=1}^{6} b_{k, d, n} v_{k, d, t-2-n} + \sum_{l=1}^{5}  b_{l, d} v_{l, d} \\\nonumber
    & + \sum_{w=1}^{167} \sum_{n=1}^{6} b_{w, d, n} x_{w, d, t-2-n} + b_{w, p_d, n} x_{w, p_d, t-2-n} + b_{w, c_d, n} x_{w, c_d, t-2-n}.
\end{align}

\begin{figure}[!ht]
\centering
\includegraphics[width=\linewidth]{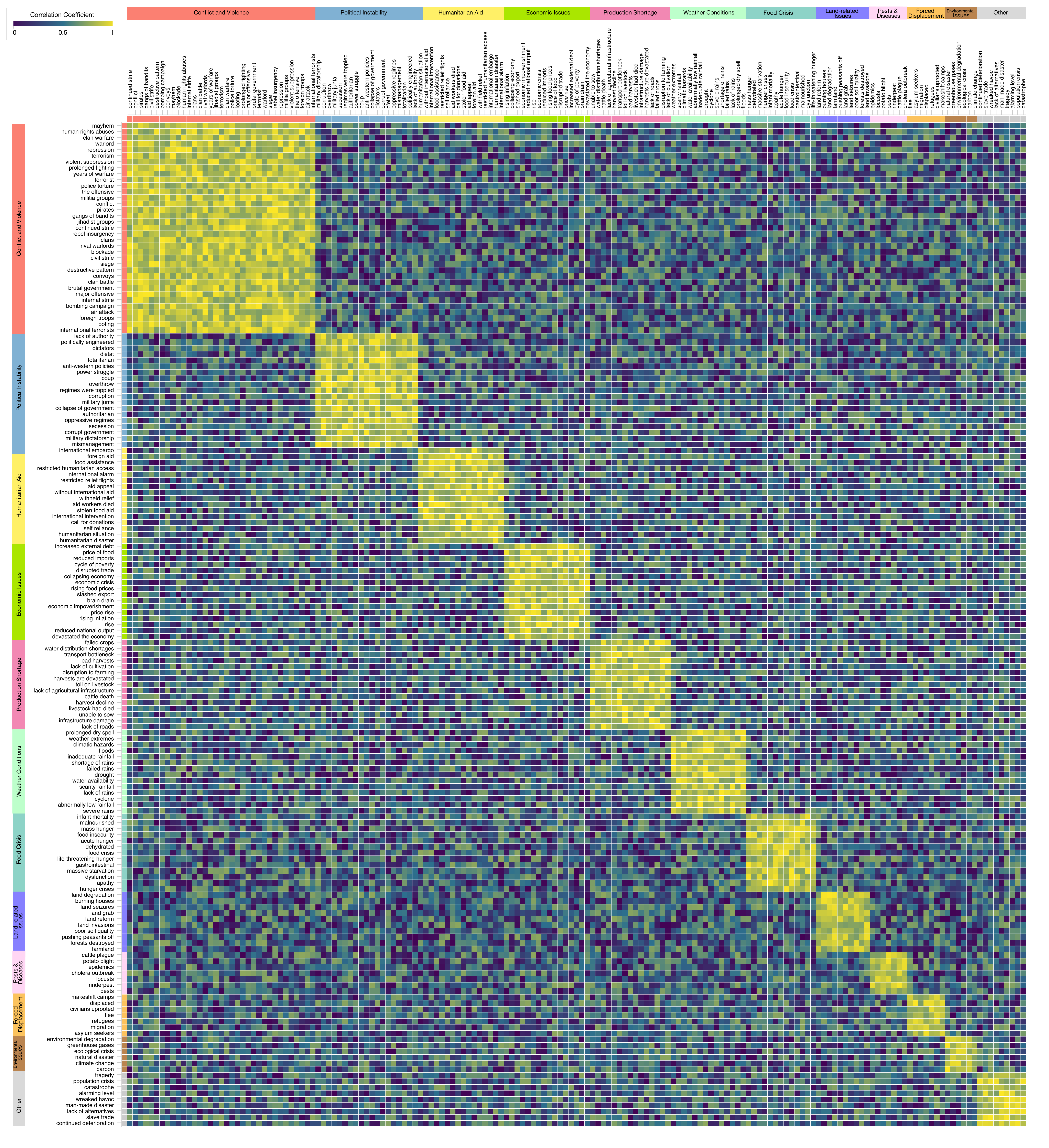}
\caption{\textbf{Clustering text features.} \footnotesize{Pairwise correlation between news factors over the period 1980-2020, showing an average correlation between news factors in the same cluster about twice as high as that of factors belonging to different clusters (69.9\% versus 34.9\%), which provides support to our choice of clustering of our text features into 12 semantically distinct clusters.}}
\label{fig8}
\end{figure}

\noindent We set: $b_{w, d, n} = b_{w, p_d, n} = b_{w, c_d, n} = 0 $ to estimate the baseline model and $b_{k, d, n} = b_{l, d} = 0 $ to estimate the news-based model. To measure each model's predictive performance, we first partition the observation period into 10 disjoint folds of equal length. We then successively train the model using a leave-one-out cross-validation strategy, ensuring that observations from each training set occur before those of the corresponding test set. We then report the average cross-validation root-mean-square-error (RMSE) across the 10 folds, both for the full model as well as for each country separately. A Lasso regularization worsens the predictions, increasing the out-of-sample RMSE by 9.9\%, 1.4\%, and 4.4\% for the baseline, news-based, and combined model respectively (Fig \ref{fig8}). 

In these estimates, each news factor is computed independently of whether a target keyword also appears in an article in which a text feature is mentioned (Fig. \ref{fig6}A and \ref{fig6}C). However, the presence of a target keyword could indicate that the text is suggesting that a food crisis is already unraveling. As a robustness check, we recompute our news factors after having excluded any article containing a target keyword. The resulting reductions in RMSE of the news-based and combined model relative to the baseline model are equal to $33.8\%$ and $39.1\%$ respectively, which represents a small deterioration of the results compared to our preferred estimates presented in Fig. \ref{fig3}.

Finally, we investigate whether the predictive performance of the model described in equation \ref{equation_model} changes by incorporating spatial averages of district-level terms accounting for the tendency of food insecurity to be spatially correlated. Let $\tilde{y}_{.,d,.}$ be the spatial average of $y_{.,d,.}$ computed using the 4 nearest neighbors of district $d$. $\tilde{x}_{.,d,.}$, $\tilde{v}_{.,d,.}$, and $\tilde{v}_{.,d}$ are defined in a similar fashion. We then estimate the following model: 

\begin{align}
\label{equation_model_spatial}
    y_{d,t} &= a_{d} + \sum_{m=1}^{6} a_{d,m} y_{d,t-3m} + \sum_{k=1}^{9} \sum_{n=1}^{6} b_{k,d,n} v_{k,d,t-2-n} + \sum_{l=1}^{5}  b_{l,d} v_{l,d} \\\nonumber
    & + \sum_{w=1}^{167} \sum_{n=1}^{6} b_{w,d,n} x_{w, d, t-2-n} + b_{w,p_d,n} x_{w,p_d,t-2-n} + b_{w,c_d,n} x_{w,c_d,t-2-n} \\\nonumber
    & + \sum_{m=1}^{6} a_{d, m} \tilde{y}_{d, t-3m} + \sum_{k=1}^{9} \sum_{n=1}^{6} b_{k,d,n} \tilde{v}_{k,d,t-2-n} + \sum_{l=1}^{5}  b_{l,d} \tilde{v}_{l,d} \\\nonumber
    & + \sum_{w=1}^{167} \sum_{n=1}^{6} b_{w,d,n} \tilde{x}_{w, d,t-2-n}.
\end{align}

\noindent We obtain an out-of-sample RMSE equal to 15\%, 9.5\%, 8.8\% for the baseline, news-based and combined model respectively (Fig. \ref{fig8}), which corresponds to a reduction in RMSE of 1.6\%, 5\% and 3.5\% respectively. Since the predictive gains are modest, we choose to keep the model more parsimonious by presenting estimates from equation \ref{equation_model} in Fig. \ref{fig3}.

\subsection{Classification of food crisis outbreaks}
\label{food_crises_classification}
We define a food crisis outbreak as a sequence of two consecutive periods during which the IPC phase raises to a value of 3 or more while the previous period's phase is smaller or equal to 2. We aim to predict an outcome variable which is equal to 1 when a food crisis outbreak occurs and zero otherwise. We convert each previously estimated model of the IPC phase into a classifier of food crisis outbreaks by introducing a lower threshold $l$ and an upper threshold $u$. An outbreak is predicted to start in district $d$ during the quarter ending on date $t$ if and only if:

\begin{alignat}{4}[left = \empheqlbrace]
   y_{d, t+1} {} \geq {} & u \\\nonumber
   y_{d, t} {} \geq {} & u \\\nonumber
   y_{d, t-1} {} \leq {} & l.\nonumber
\end{alignat}

\noindent Each model's thresholds $l$ and $u$ determine its precision and its recall. By varying $l$ and $u$ from $1$ to $5$ in increments of size $0.1$, we can estimate a model's Pareto front. We then fix all the models' precision to be equal to 80\% and we compare their recall values measured on the Pareto front. We obtain threshold values for $(l,u)$ equal to $(2.2, 3.1)$, $(1.9, 2.7)$, and $(2.1, 3.3)$ for the baseline, news-based, and combined model respectively. While we are agnostic about how to balance type I and type II errors, our results show an improvement along the Precision-Recall curves across countries (Fig \ref{fig11}). 

At the time of publishing the IPC phase, FEWS NET additionally provides a projection of next period's values (“expert model”). In line with our previous analysis, we binarize these expert forecasts to produce a predictive model of food crisis outbreaks in which an outbreak occurs when the IPC phase raises to 3 or more for at least least two consecutive periods. The expert model's precision and recall are equal to 70\% and 66\% respectively, which represents a degradation compared to the combined model for which we obtained a precision and a recall respectively equal to 80\% and 86\%. 

\subsection{Statistics}
In Fig. \ref{fig3}, we report the average RMSE of each model across 10-fold cross validation, ensuring that observations used for training occur prior to those included in the test set.

\subsection{Data availability}
The data that support the findings of this study are available on Dataverse or from the corresponding author upon reasonable request.

\subsection{Code availability}
The data that support the findings of this study are available on GitHub or from the corresponding author upon reasonable request.

\section*{Acknowledgments}
We thank Alice Grishchenko for outstanding visual design work, Bo Andree and Andres Chamorro for sharing their data, and Bo Andree, Delia Baldassarri, Nir Grinberg, Aart Kray, David Lazer, Rohini Pande, Nadia Piffaretti, and Damien Puy for useful discussions and feedback. This study was supported by the World Development Report 2021: Data for Better Lives (S.F.). The findings, interpretations, and conclusions expressed in this paper are entirely those of the authors. They do not necessarily represent the views of the International Bank for Reconstruction and Development/World Bank and its affiliated organizations, or those of the Executive Directors of the World Bank or the governments they represent.

\section*{Author Contributions}
Conceptualization: AB, LS, SF.
Methodology: AB, LS, SF.
Investigation: AB, LS, SF.
Visualization: AB, LS, SF.
Funding acquisition: LS, SF.
Project administration: AB, LS, SF.
Supervision: AB, LS, SF.
Writing: AB, LS, SF.

\section*{Competing Interests and Conflict of Interests Disclosures}
Dr. Fraiberger reports receipt of grant funding from the World Bank World Development Report 2021: Data for better lives. Dr. Subramanian is a co-founder of Entrupy Inc, Velai Inc, and Gaius Networks Inc and has served as a consultant for the World Bank and the Governance Lab. Mr. Balashankar is a Ph.D student at New York University, and is also funded in part, by the Google Student Research Advising Program. Dr. Subramanian reports that Velai Inc broadly works in the area of socio-economic predictive models, and Mr. Balashankar holds a small percentage of equity in Velai Inc, for licensed technology through New York University. No other disclosures were reported.

\section*{Materials and correspondence}
Please contact Samuel Fraiberger (sfraiberger@worldbank.org) for correspondence related to material requests.

\section*{Extended Data}

\begin{figure}[!ht]
\centering
\includegraphics[width=\linewidth]{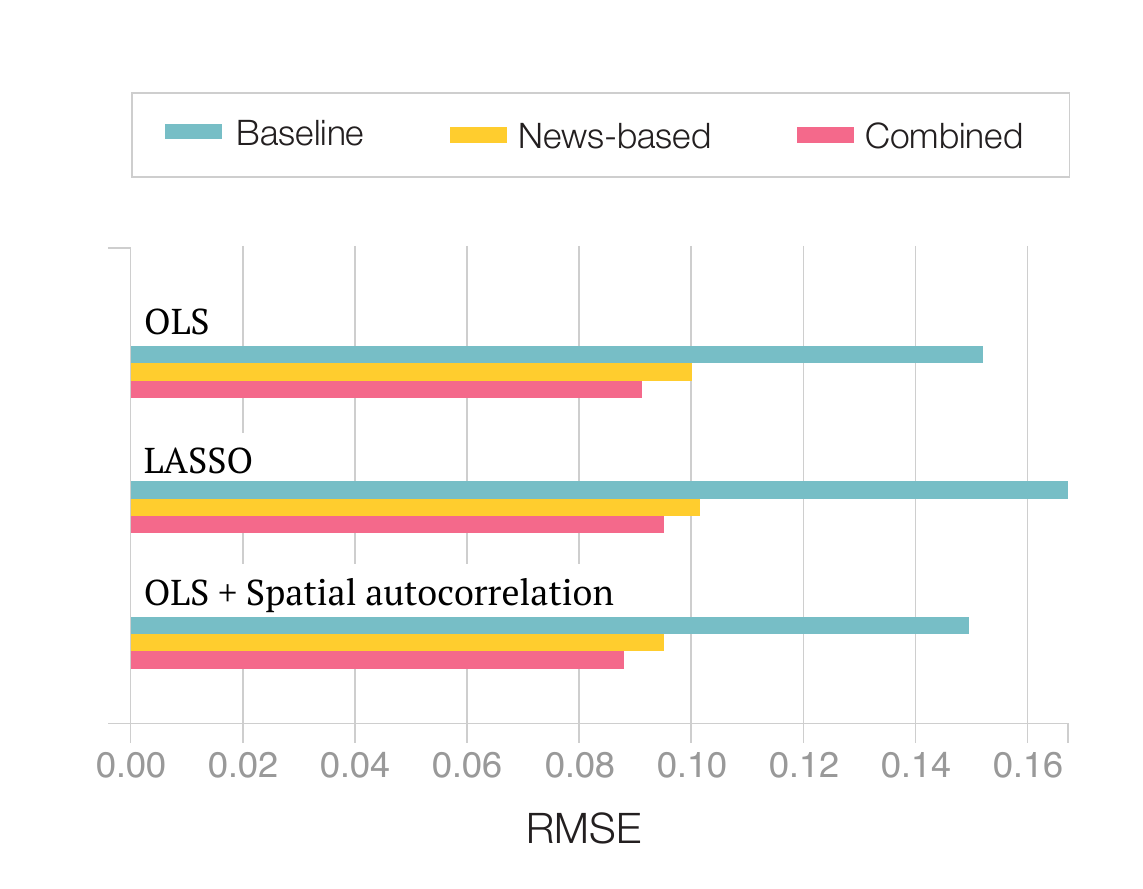}
\caption{\textbf{Alternative specifications.} \footnotesize{We first compare the OLS estimates of equation \ref{equation_model} shown in Fig.\ref{fig3}A with estimates of the same model using lasso regularization, showing that it leads to a degradation of the out-of-sample RMSE. We then demonstrate that the model described by equation \ref{equation_model_spatial} which incorporates spatial averages of district-level terms leads to a small reduction of the out-of-sample RMSE. Since the predictive gains are modest, we choose equation \ref{equation_model} as our main specification to keep the model more parsimonious.}}
\label{fig9}
\end{figure}

\begin{figure}[!ht]
\centering
\includegraphics[width=\linewidth]{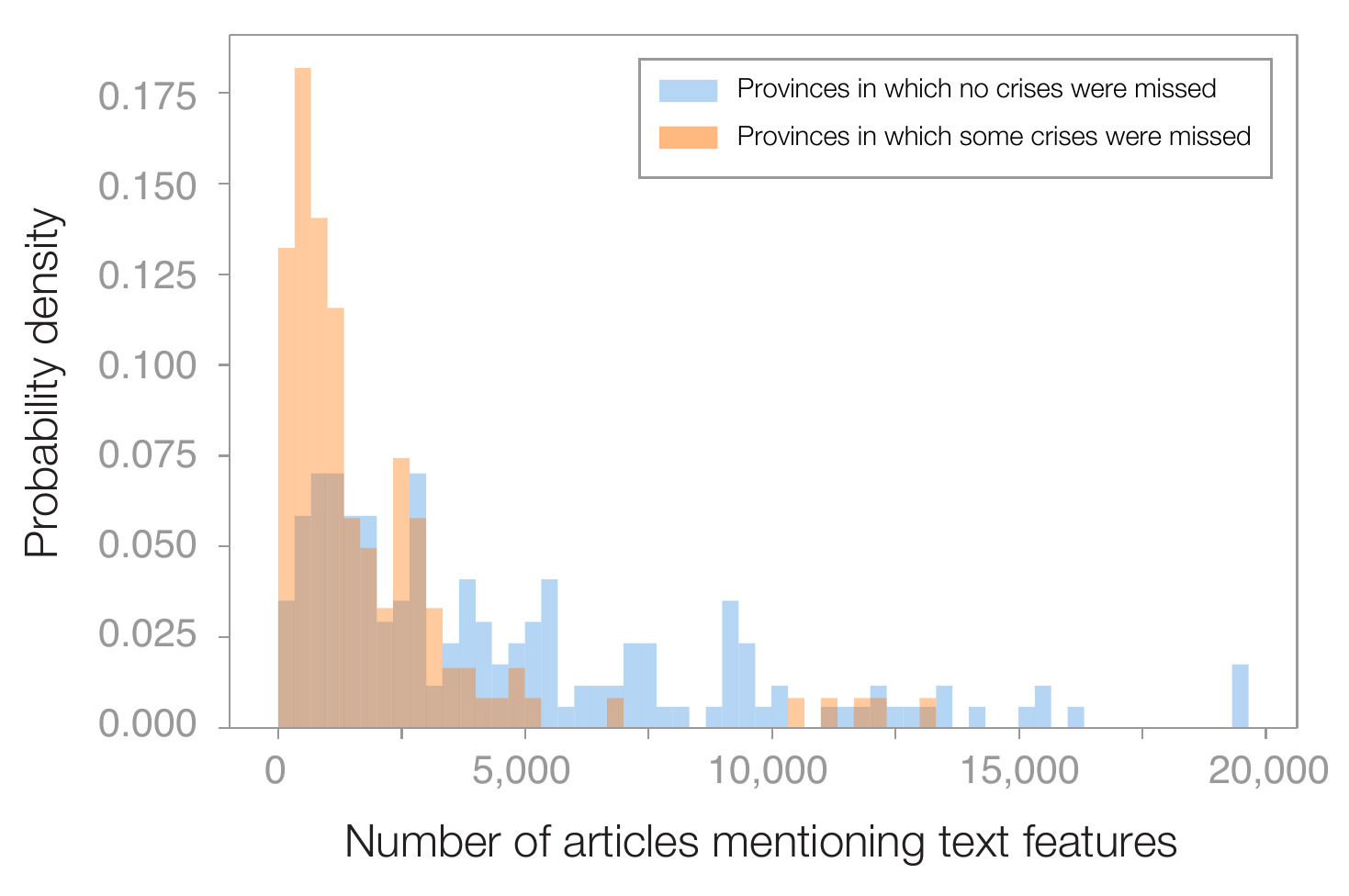}
\caption{\textbf{News coverage and predictive performance.} \footnotesize{Distribution of the number of news articles mentioning text features across administrative units of level 1 (“provinces”), separating between provinces in which the combined model predicts all the crisis outbreaks (blue) from those in which it fails to predict at least one crisis (orange), which reveals that provinces in which the combined model fails to predict some crisis outbreaks have lower news coverage that those in which the model predicts all of them.}}
\label{fig10}
\end{figure}

\begin{figure}[!ht]
\centering
\includegraphics[width=\linewidth]{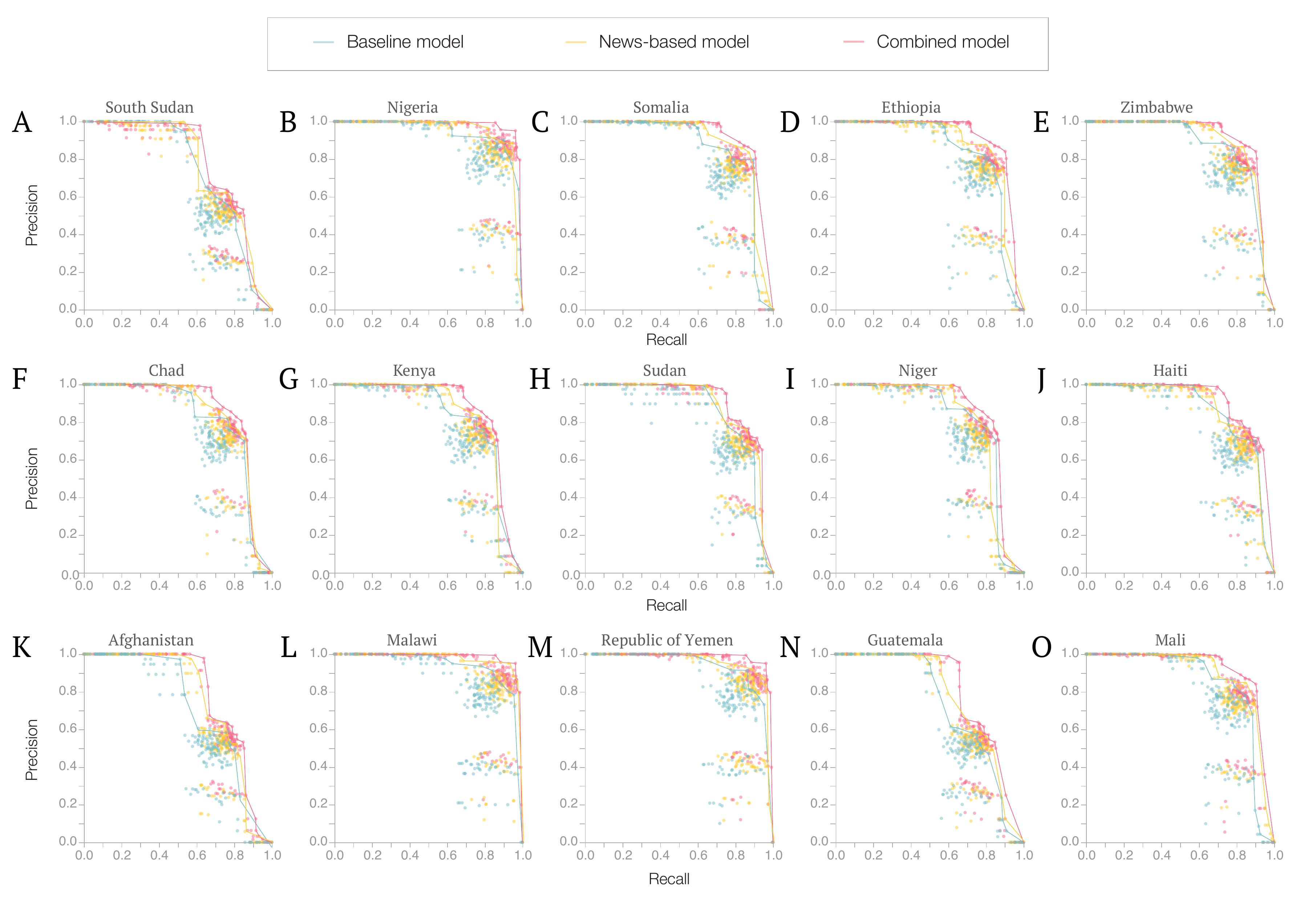}
\caption{\textbf{Precision-Recall curves.} \footnotesize{We show the same precision-recall curves as the one described in Fig.\ref{fig3}B, after having split the evaluation set by country, which indicates that the combined model also outperforms both the news-based and the baseline model at the country level.}}
\label{fig11}
\end{figure}

\begin{figure}[!ht]
\centering
\includegraphics[width=\linewidth]{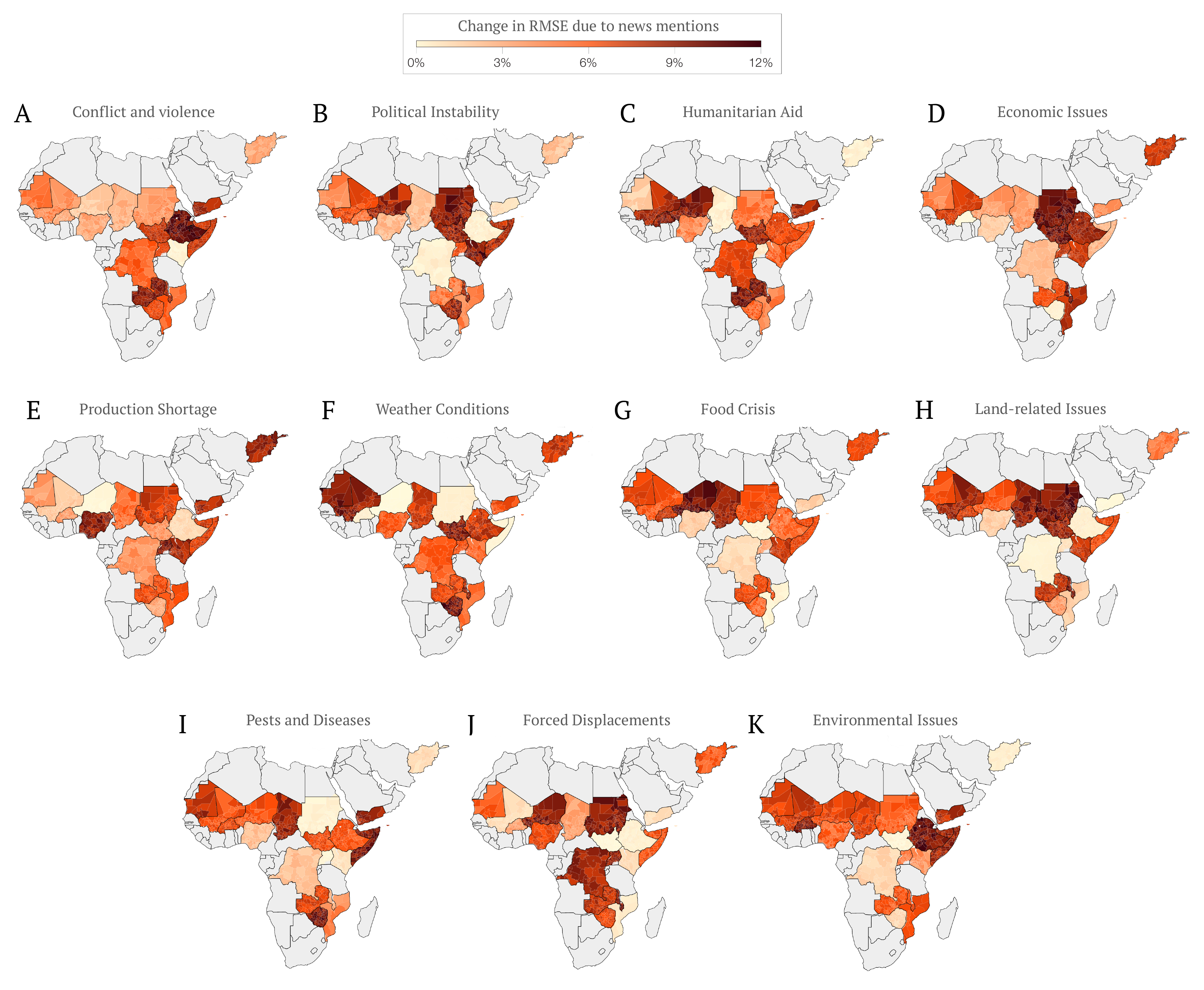}
\caption{\textbf{Ablated models.} \footnotesize{We re-estimate the combined model by removing each cluster of news factors (“ablated model”). We report the district-level increase in RMSE of each ablated model relative to the combined model (A-K), allowing us to identify the regions in which each cluster of news factors provides the highest contribution to the prediction.}}
\label{fig12}
\end{figure}

\end{document}